# Time Series Forecasting of HIV/AIDS in the Philippines Using Deep Learning: Does COVID-19 Epidemic Matter?

Sales G. Aribe Jr.[1], Bobby D. Gerardo[2], Ruji P. Medina[3]

[1,2,3]Technological Institute of the Philippines, Quezon City, Philippines
[1]Information Technology Department, Bukidnon State University, Malaybalay City, Philippines
[2]Northern Iloilo State University, Estancia, Iloilo, Philippines

*Abstract* – With a 676% growth rate in HIV incidence between 2010 and 2021, the HIV/AIDS epidemic in the Philippines is the one that is spreading the quickest in the western Pacific. Although the full effects of COVID-19 on HIV services and development are still unknown, it is predicted that such disruptions could lead to a significant increase in HIV casualties. Therefore, the nation needs some modeling and forecasting techniques to foresee the spread pattern and enhance the government's prevention, treatment, testing, and care program. In this study, the researcher uses Multilayer Perceptron Neural Network to forecast time series during the period when the COVID-19 pandemic strikes the nation, using statistics taken from the HIV/AIDS and ART Registry of the Philippines. After training, validation, and testing of data, the study finds that the predicted cumulative cases in the nation by 2030 will reach 145,273. Additionally, there is very little difference between observed and anticipated HIV epidemic levels, as evidenced by reduced RMSE, MAE, and MAPE values as well as a greater coefficient of determination. Further research revealed that the Philippines seems far from achieving Sustainable Development Goal 3 of Project 2030 due to an increase in the nation's rate of new HIV infections. Despite the detrimental effects of COVID-19 spread on HIV/AIDS efforts nationwide, the Philippine government, under the Marcos administration, must continue to adhere to the United Nations' 90-90-90 targets by enhancing its ART program and ensuring that all vital health services are readily accessible and available.

## I. Introduction

While the whole world is waging war against the invisible enemy, COVID-19, 2021 commemorates 40 years of the first detection of the first Human Immunodeficiency Virus (HIV) in the world[1] and 38 years for the Philippines.

Since then, 95,212 total diagnosed cases in the country have been documented by the HIV/AIDS and ART Registry of the Philippines (HARP) as of January 2022[2]. Attempts to eradicate HIV transmission have significantly reduced new HIV infections, HIV-related fatalities, and costs for antiretroviral therapies (ART)[3]. Regrettably, the problem of HIV in the Philippines stays remarkably high given the scarcity of HIV prevention and treatment programs, testing services, and counseling[4]. This problem makes it challenging to convince patients to screen and establish high levels of ART compliance and care maintenance. It is also attributed to a lack of awareness, obstacles to obtaining care due to COVID-19 restrictions, and widespread stigma[5]. Additionally, high incidence of therapeutic failure, lack of follow-up, and inadequate medical coordination have undesirable effects on several HIV patients[6, 7].

Forecasting epidemics like HIV and modeling its outcome critically impact health systems and policymakers[8, 9]. To recommend new public health measures and strategies and assess the efficiency of current regulations, policymaking relies on judgments generated by forecasting models[10]. To deliver the best treatments and protective and preventive measures, the governments and health professionals will benefit from knowing the trajectory of the epidemic's future spread[11]. Using Artificial Intelligence (AI) techniques to predict the epidemic's spread is a viable replacement for prevalent epidemic models[12].

Data sequences gathered over time are often utilized as input data for machine learning to determine the course of the HIV epidemic. Various commonly used approaches have been employed to anticipate the spread pattern of HIV.

For example, when estimating the initial growth dynamics of the HIV outbreak in Brazil, the Generalized Growth Model (GGM) was utilized to calculate the scaling of growth characteristics with measurable probability[13]. However, the model relies on AIDS surveillance data, and the AIDS cases were underreported at 42.7%. The Modes of Transmission (MoT) technique also mapped HIV exposures at Morocco's national and local levels[14]. Nevertheless, it has a static model structure and does not consider variability in the risk of HIV contamination. Table 1 displays several noteworthy and current AI and statistical applications for predicting HIV epidemics.

TABLE 1
ADDITIONAL AI AND MATHEMATICAL APPLICATIONS FOR PREDICTING HIV/AIDS

| Methods | Region | Findings | Ref. |
|---|---|---|---|
| Multiple Methods: LSTM, ARIMA, GRNN, and ES | Guangxi, China | *Pros*: The LSTM model's MSE was the least when the time step was 12. *Cons*: ES and GRNN models did poorly in forecasting HIV incidence in Guangxi, China. | [15] |
| Autoregressive Integrated Moving Average (ARIMA) | Indonesia, Finland, Nigeria, Korea, Brazil, Uganda, & Various Countries | *Pros*: The model is capable of producing a graphically dynamic future prediction. *Cons*: The MAE equivalent needs further improvement. | [16], [17], [18], [19], [20], [21], [22] |
| Box-Jenkins | Philippines | *Pros*: The series is a good fit for the model. *Cons*: Most ACF and PACF results are within acceptable limits but unremarkable. | [23] |
| Box-Jenkins using Catch-All Technique | Zimbabwe | *Pros*: In general, the data are steady and reliable for predicting new HIV infections. *Cons*: The study focused only on a local level. | [24] |
| Univariate Box-Jenkins | Philippines | *Pros*: SARIMA (2,1,0)(0,0,1) is the suitable model. *Cons*: Use of optimization such as differential evolution. | [25] |

There has been a substantial increase in research activity in the past ten years due to interest in applying artificial neural networks (ANN) for predicting[26]. The ANN has demonstrated considerable potential in predicting better values for precision and accuracy among the various AI algorithms[8]. The *artificial neurons* that make up this model's network of interconnected nodes serve as a loose representation of the neurons found in a biological brain. We may train the network to deal with various problems by implementing this algorithm that mimics the workings of real neurons [27]. In many applications, it would be the most well-known and typical network model in predicting various disciplines [26], ranging from stock price, electric load, weather prediction, oil price, plant classification, air quality index forecasting, and many others[28, 29]. Additionally, it is generally acknowledged for epidemiologic analysis of illnesses' risk and epidemic size to estimate the incidence of diseases using the ANN algorithm[30]. Table 2 illustrates how the ANN algorithm forecasts HIV/AIDS epidemics.

TABLE 2
HIV/AIDS TIME SERIES FORECASTING USING ANN MODEL

| Country | Scope of Study | Ref. |
|---|---|---|
| China | Statistics on the prevalence of HIV in Guangxi, China, from 2005 to 2016. | [15] |
| Nigeria | Occurrence and spread of HIV/AIDS in Niger Delta, Nigeria. | [31] |
| Egypt | ART Coverage in Egypt for the period 2000-2018. | [32] |
| China | Evaluation and analysis of the incidence of AIDS in China employing the BP ANN and ARIMA. | [33] |
| Kenya | The MLP's (Multilayer Perceptron) ART implementation in Kenya from 2000 to 2018. | [34] |
| South Africa | From 2000 to 2018, MLP was used to study the impact of ART and its penetration in South Africa. | [35] |
| Zimbabwe | Pregnant women at GDH, Zimbabwe, have new HIV transmissions. | [36] |
| Malawi | ART coverage in Malawi from 2000 to 2018 using the MLP-ANN model. | [37] |
| Philippines | Impact of ART in the Philippines employing ANN from 2009 to 2022 (156 months). | [9] |

It should be noted that no ANN model uses the COVID-19 dataset (period 2020–2022) to predict the HIV transmission pattern in the future, neither in the Philippines nor globally.

The ANN algorithm has been used to anticipate the monthly incidence of HIV in the Philippines as part of a fast-track approach to stop the AIDS pandemic by 2030 based on the third Sustainable Development Goal (SDG-3), also known as Project 2030. Using the COVID-19 dataset in the country, which starts in January 2020, the paper will study the impact of coronavirus on the new HIV incidence rate by evaluating the country's progress toward a 90 percent decrease between 2010 and 2030 [38]. Therefore, this study investigates the interactions between the HIV and COVID-19 epidemics, especially the impact of the COVID-19 outbreak on HIV response [39].

The main goal of this study is to develop an ANN model that can be used to forecast the monthly incidence of HIV in the Philippines. In particular, it aims to:

1. Determine whether the COVID-19 dataset and the ANN model are a perfect fit for forecasting;
2. Predict the monthly and total number of cases through the end of 2030;
3. Find out the Philippines' progress in the implementation of Project 2030's SDG-3;
4. Compare the results of the monthly HIV infections that are forecasted against the actual cases during the COVID-19 outbreak in the Philippines; and
5. Determine the accuracy rate of the ANN model using the commonly used statistical method for performance measurements.

## II. MATERIALS AND METHODS

### A. Data Source

COVID-19 dataset from January 2020 to February 2022 was used to forecast HIV spread in the country. These ART data, which contain all confirmed HIV-positive individuals, are collected from the Philippine Department of Health (DOH) website, specifically from the Epidemiology Bureau (EB). The official registry of all diagnoses and other ART information is extracted from the San Lazaro Hospital STD/AIDS Cooperative Central Laboratory (SACCL) and documented by HARP. These data, registered in the system and receiving ARV therapies throughout the period, are provided by 158 primary HIV healthcare and treatment centers in the Philippines.

The historical HIV/AIDS cases in the Philippines during the corona virus outbreak are shown in Table 3 and illustrated in Figure 1 to show pandemic trends. In 1984, when AIDS was discovered in the Philippines, records of cumulative cases were kept by HARP.

TABLE 3
MONTHLY CASES OF HIV IN THE PHILIPPINES DURING COVID-19

| Index | Period (Month & Year) | Monthly Cases | Aggregated Cases |
|---|---|---|---|
| 1 | Jan-20 | 1,039 | 75,846 |
| 2 | Feb-20 | 1,227 | 77,073 |
| 3 | Mar-20 | 552 | 77,625 |
| 4 | Apr-20 | 257 | 77,882 |
| 5 | May-20 | 187 | 78,069 |
| 6 | Jun-20 | 490 | 78,559 |
| 7 | Jul-20 | 523 | 79,082 |
| 8 | Aug-20 | 133 | 79,215 |
| 9 | Sep-20 | 1,219 | 80,434 |
| 10 | Oct-20 | 735 | 81,169 |
| 11 | Nov-20 | 620 | 81,789 |
| 12 | Dec-20 | 1,076 | 82,865 |
| 13 | Jan-21 | 888 | 83,753 |
| 14 | Feb-21 | 855 | 84,608 |
| 15 | Mar-21 | 1,038 | 85,646 |
| 16 | Apr-21 | 1,119 | 86,765 |
| 17 | May-21 | 645 | 87,410 |
| 18 | Jun-21 | 1,495 | 88,905 |
| 19 | Jul-21 | 1,045 | 89,950 |
| 20 | Aug-21 | 878 | 90,828 |
| 21 | Sep-21 | 981 | 91,809 |
| 22 | Oct-21 | 1,136 | 92,945 |
| 23 | Nov-21 | 1,268 | 94,213 |
| 24 | Dec-21 | 993 | 95,206 |
| 25 | Jan-22 | 875 | 96,081 |
| 26 | Feb-22 | 1,054 | 97,135 |

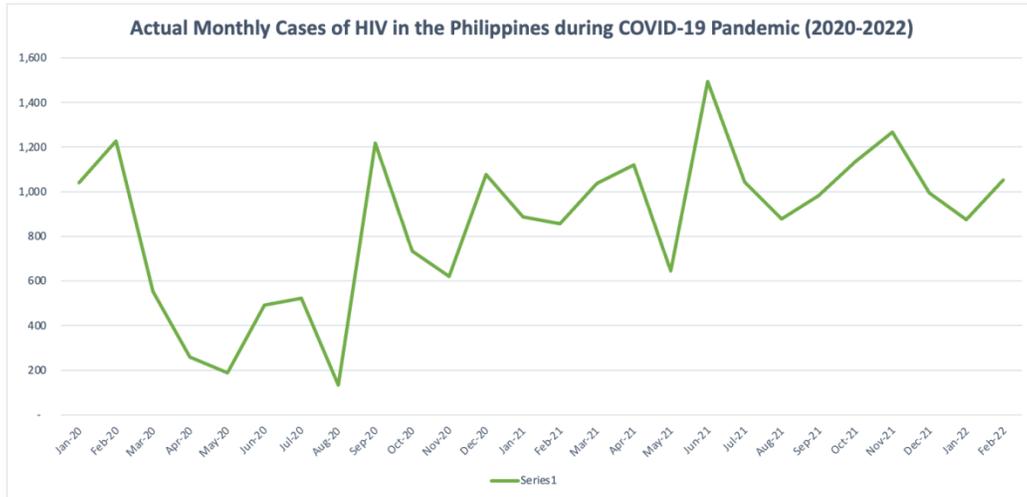

Fig. 1 HIV Epidemic in the Philippines During COVID-19

*B. Conceptual Framework*

The ANN algorithm was used to estimate monthly cases of HIV in the Philippines up to the month of December 2030. Figure 2 presents the detailed procedures carried out to conduct this study. HIV/AIDS cases at the start of the COVID-19 outbreak in the nation are meticulously analyzed and organized sequentially using MATLAB software during the data pre-processing stage. Based on the parameters and variables of the problem under investigation, the ANN design for the outbreak forecast is problem-dependent. For forecasting using the neural net time series tool (*ntstool*), the network design would be affected by the quantity of input, hidden, and output nodes. The choice of structure can strongly influence similar information amongst variables[40]. The HIV data was used for training, validation, and testing to estimate the new HIV incidence in the Philippines. Standard metrics for regression were used for error and accuracy analysis between observed and predicted HIV/AIDS cases.

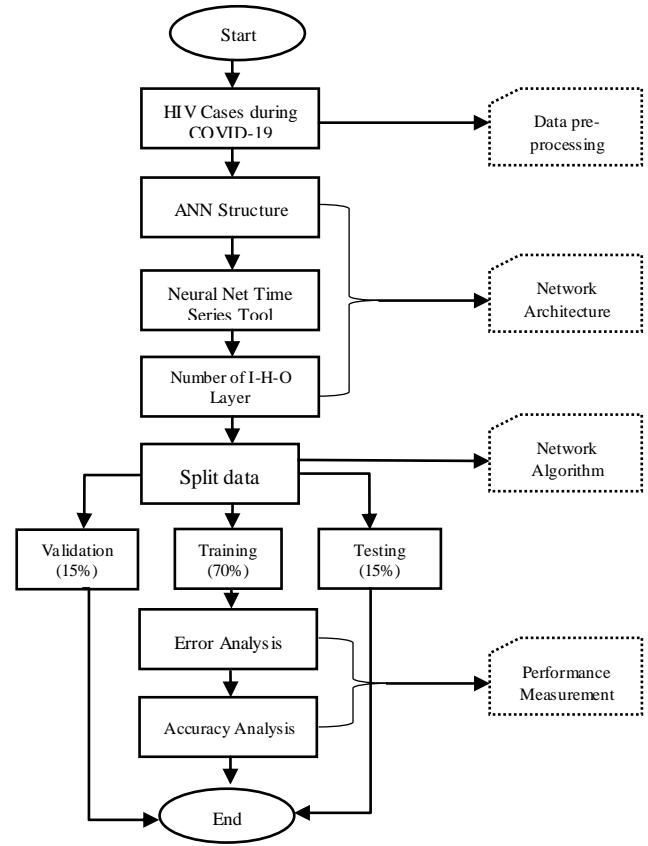

Fig. 2 ANN Model's Conceptual Framework in Forecasting HIV Cases

*C. Artificial Neural Network*

The forecasting capability of ANN is utilized in this study to forecast the HIV's spread pattern. It is an effective technique for handling issues with statistical evolution that determines how to organize the way signals are transmitted in the biological neural network. This flexible processing paradigm creates output based on nonlinear interactions between parameters and variables and takes any input[41]. Like regression, ANNs are intended to comprehend the relationship between input and output variables. Time series analysis, which may be used to estimate future values utilizing historical data, is the most popular method of using ANNs for prediction[42]. This model is capable of plotting input and output. It functions as a multi and parallel processor and distributed system when the inputs and weights flow through the suitable neurons[43].

Figure 3 shows a typical ANN. It consists of three parts: a collection of connection links or synapses that pass inputs $x$ multiplied by the weight, $w_i x_i$, a summing function $S$, and the activation function $\delta(x_i)$ for restricting the output of a neuron's amplitude. In figure 2, $w_1, w_2 \ldots w_i$, are the group of weights, and $x_1, x_2 \ldots x_i$, are the group of inputs to the neuron. Equations 1 through 4 express inputs after weight multiplication:

$$S = w_1 x_1 + w_2 x_2 + \cdots next, = \sum w_i x_i \quad (1)$$

$$S = \sum_{i=1}^{n}(w_i x_i - \theta) \quad (2)$$

$$S = f(x) = x \quad (3)$$

$$\delta = b - s = b - x \quad (4)$$

where $\theta$ = is a bias term and $i$ = are numbers from 1 to n.

The threshold value must be determined using the foregoing equations. Because there is only one input and one output, a neuron with the proper weight and threshold value returns a value in time units. The threshold process is crucial as it determines how much input and output differ from one another. If the weight of the output sum $S(\delta(x_i))$ is above than this threshold level, the outcome will be 1, else, it will be 0. One way of putting it is as follows:

$$\delta(x_i) = \begin{cases} 1 & x_i \geq 0 \\ 0 & x_i < 0 \end{cases} \quad (5)$$

There are typically four threshold functions: sigmoid, signum, piecewise linear, and hyperbolic tangent [45]. In the current work, input and hidden layer neurons were associated using a tangent sigmoid function (*tanh*), and hidden and output layer neurons were implemented using a linear transfer function. In neural networks, where speed and precision are essential, this function has an optimal choice, and the transfer function's exact value is accurate[46].

The MLP neural network, the most well-known and extensively applied neural network design, may help structure future intelligence by enabling perception and learning ability[47]. In the current study, a nonlinear autoregressive (NAR) method was used together with a multilayer feed-forward neural network structure.

Estimated future time series values are predicted using the time series' historical values, a dynamic filtering type. In neural networks, tap-delay lines are utilized for nonlinear filtering and forecasting. The NAR tool addresses time series issues in *ntstool* using the MATLAB R2021a version. Using the preceding number of *y(t)*, it estimates the time series value of *y(t)*. In a mathematical formula, it can be written as:

$$yt = f(y(t-1) + y(t-2) + y(t-3) + y(t-m)) \quad (6)$$

The MLP model was utilized, although overfitting or underfitting is the fundamental problem with this method. As a result, there is a disparity between the input and output layers. The *Stop Training Approach* has been used to reduce these inaccuracies drastically. Data on monthly HIV cases are gathered and randomly divided into training, testing, and validation data.

Another critical step affecting the output's accuracy is the choice of the hidden layer. Hidden layers and feedback delays are assigned to 10 to achieve the desired value for this model.

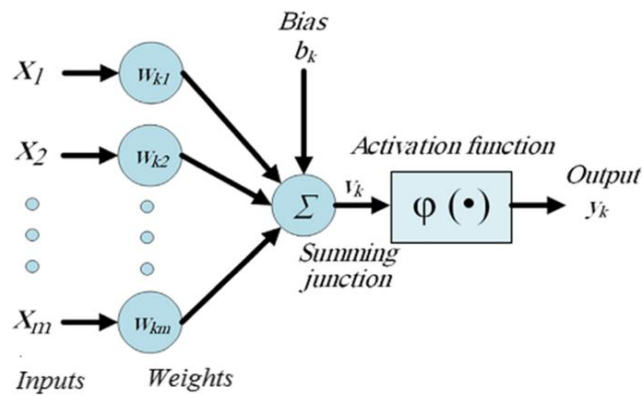

**Fig. 3 Basic Artificial Neural Network[44]**

*D. Training of the Model*

Data from 26 months which makes up about 70%, were trained for predicting monthly incidences of the HIV outbreak in the Philippines. These data were automatically trained and processed using MATLAB software to determine the desired output *y(t)*. Due to limited datasets provided by monthly data, values are converted to weekly data using the Microsoft Excel Date and Time Function. Thus, 114 weekly data was produced during the conversion process. 70% of training data equivalent to 80 HIV weekly cases time series data is divided into a $1 \times 80$ cell array of $1 \times 1$ matrix. In this instance, training is associated with the learning process of ANN. Testing and validation (15% each) are employed from 30% of the 114 data. This ratio suggests that the remaining 34 HIV weekly cases time series data are divided into a $1 \times 34$ cell array of $1 \times 1$ matrix. The selection ratio is automatically adjusted, and the value of training is affected when the testing and validation are altered from 5% to 35%. Nevertheless, the recommended choice and default ratio is 70%: 15%: 15%. Forecasted weekly data will be converted to monthly data once training is done.

*E. Evaluation Metrics*

As forecasting produces varied data, error observations are essential for the forecasting process. Hence, this study uses the following metrics of regression, as shown in Equations 7–10, to evaluate the performance of the ANN forecasting model and for after-the-fact error analysis[48]:

$$RMSE = \sqrt{\frac{1}{N}\sum_{i=1}^{N_s}(P_{yi} - Y_i)^2} \quad (7)$$

$$MAE = \frac{1}{N}\sum_{i=1}^{N_s}|P_{yi} - Y_i| \quad (8)$$

$$MAPE = \frac{1}{N}\sum_{i=1}^{N}|\frac{P_{yi} - Y_i}{YP_i}| \quad (9)$$

$$R^2 = 1 - \frac{\sum_{i=1}^{n}(Y_i - P_{yi})^2}{\sum_{i=1}^{n}(Y_i - \bar{Y}_i)^2} \quad (10)$$

The best approach is the one with the lowest RMSE, MAPE, and MAE values. A higher $R^2$ value denotes a better and more favorable correlation for any method[49].

### III. RESULTS AND DISCUSSIONS

*A. On the COVID Dataset's Validity as a Perfect Fit for the ANN Model in Prediction*

Numerous structures have been verified by adjusting the neurons of hidden layers throughout the training, validation, and testing stages. Due to its low RMSE and good $R^2$ values, the ANN structure 1-10-1, representing Input, Hidden, and Output, was touted as the best structure. When applied to the recommended model, the dataset covering the COVID-19 pandemic in the Philippines has been found applicable to forecast future monthly HIV/AIDS cases. Thus, the R-value of the entire data is 0.93754, while the trained and tested model is 0.95366 and 0.90746, respectively, which indicates a very high prediction accuracy, as shown in Figure 4 on the regression plot of the model. The figure 5 displays the corresponding autocorrelation plot of the model. The values are clearly moving in the direction of 0, which indicates a stronger favorable correlation.

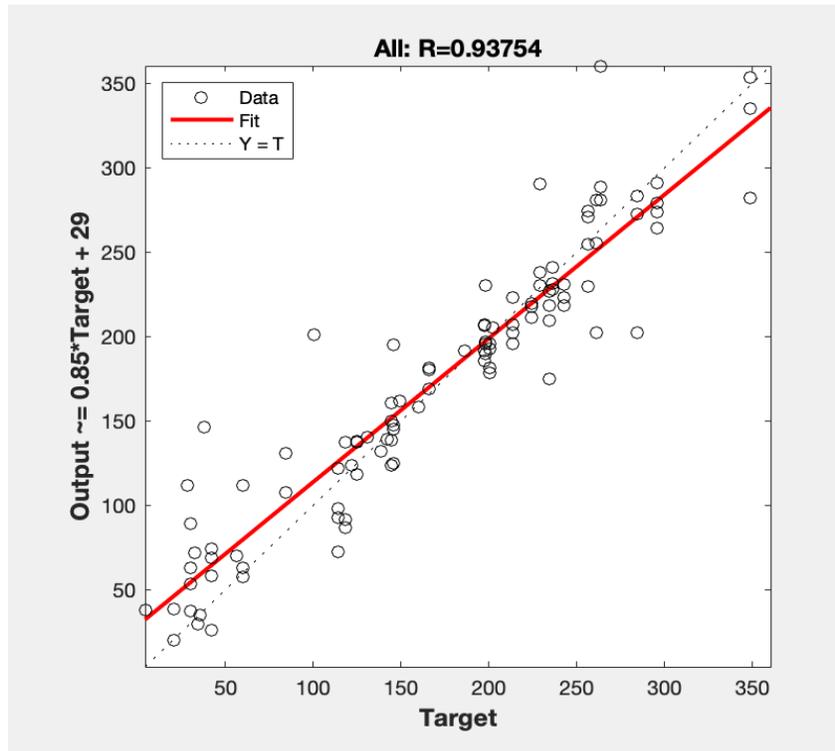

**Fig. 4 Regression Coefficient (R=0.93754) with 10 Hidden Layer Neurons of all Dataset**

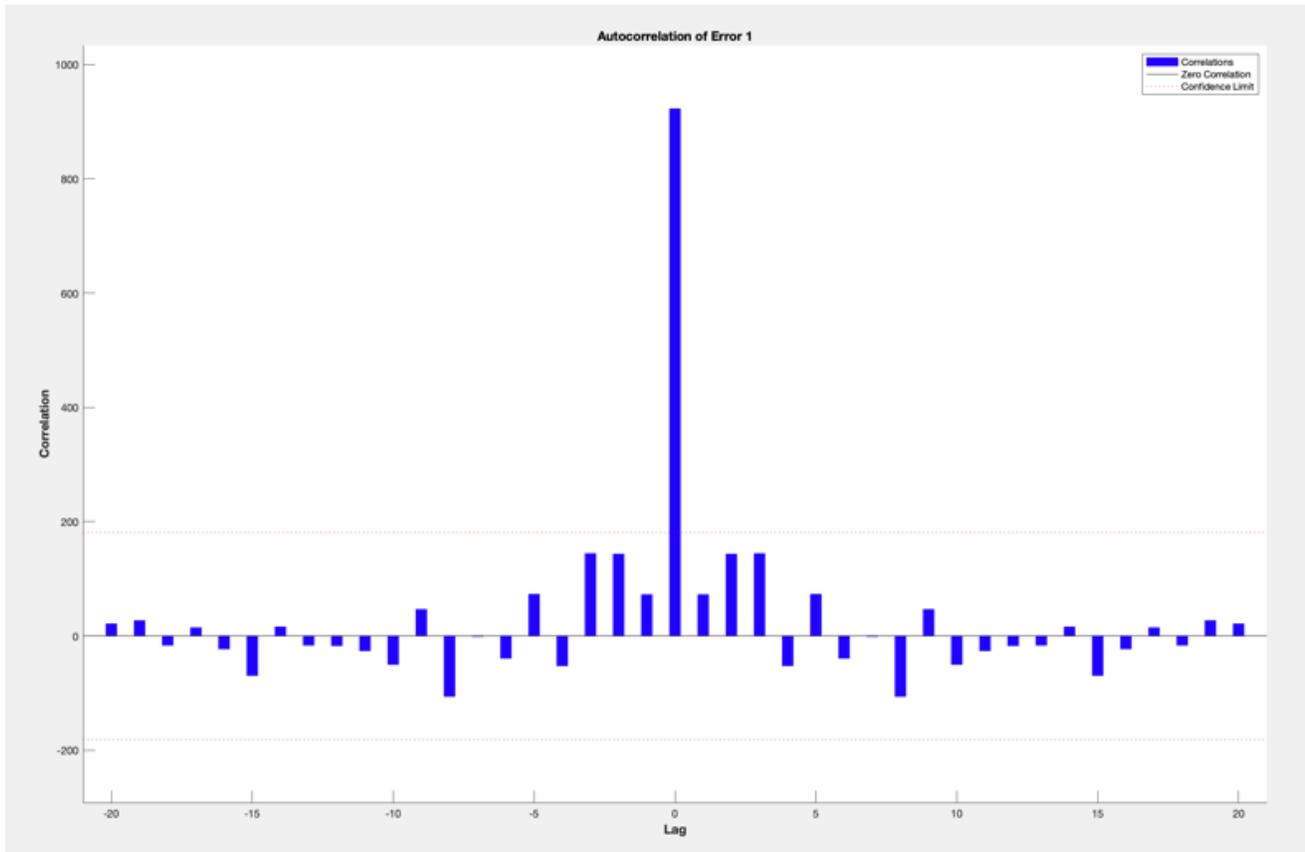

**Fig. 5 Autocorrelation Plot of the Trained, Validated, and Tested Model**

*B. On the Estimated Monthly and Aggregated Cases of HIV Until the Year 2030*

Using Table 4, the number of monthly and aggregated HIV cases in the Philippines covering months 27-158 (March 2022 – December 2030) can be predicted using the ANN model. It is estimated that the number of aggregated cases of infection by December 2030 will be 145,723, about a 2-fold growth from February 2022 actual cases of 93,557.

Nevertheless, it is projected that by 2030, the number of cases per month will only be 345, a significant decline from the actual number of cases in February 2022 of 1,054. Moreover, it is clear from the estimated cases that March is the typical month for the seasonal maxima. The maximum number of new HIV infections is foreseen in March 2022, with 1,772 cases, while the lowest new HIV incidence is in December 2023, with 171 cases.

TABLE 4
SAMPLE FORECAST OF HIV CASES IN THE PHILIPPINES (JANUARY 2029 – DECEMBER 2030)

| Period (Month & Year) | Monthly Cases | Aggregated Cases |
|---|---|---|
| Jan-29 | 375 | 130,002 |
| Feb-29 | 373 | 130,375 |
| Mar-29 | 223 | 138,349 |
| Apr-29 | 457 | 138,806 |
| May-29 | 396 | 139,202 |
| Jun-29 | 261 | 139,463 |
| Jul-29 | 355 | 139,818 |
| Aug-29 | 386 | 140,204 |
| Sep-29 | 304 | 140,507 |
| Oct-29 | 338 | 140,845 |
| Nov-29 | 326 | 141,171 |
| Dec-29 | 407 | 141,579 |
| Jan-30 | 318 | 141,896 |
| Feb-30 | 302 | 142,198 |
| Mar-30 | 271 | 142,468 |
| Apr-30 | 437 | 142,906 |
| May-30 | 339 | 143,245 |
| Jun-30 | 238 | 143,482 |
| Jul-30 | 420 | 143,902 |
| Aug-30 | 360 | 144,262 |
| Sep-30 | 361 | 144,623 |
| Oct-30 | 316 | 144,939 |
| Nov-30 | 327 | 145,265 |
| Dec-30 | 457 | 145,723 |

It is interesting to note that, according to Figure 6, the anticipated values display random cycles, albeit with downward veering in the following years. With abrupt leaps, the amplitude or distance lengthens. On the other hand, Figure 7 shows an upward linear trend, indicating that HIV occurrences in the Philippines will continue to climb. MATLAB software automatically generates these charts or figures after training the data.

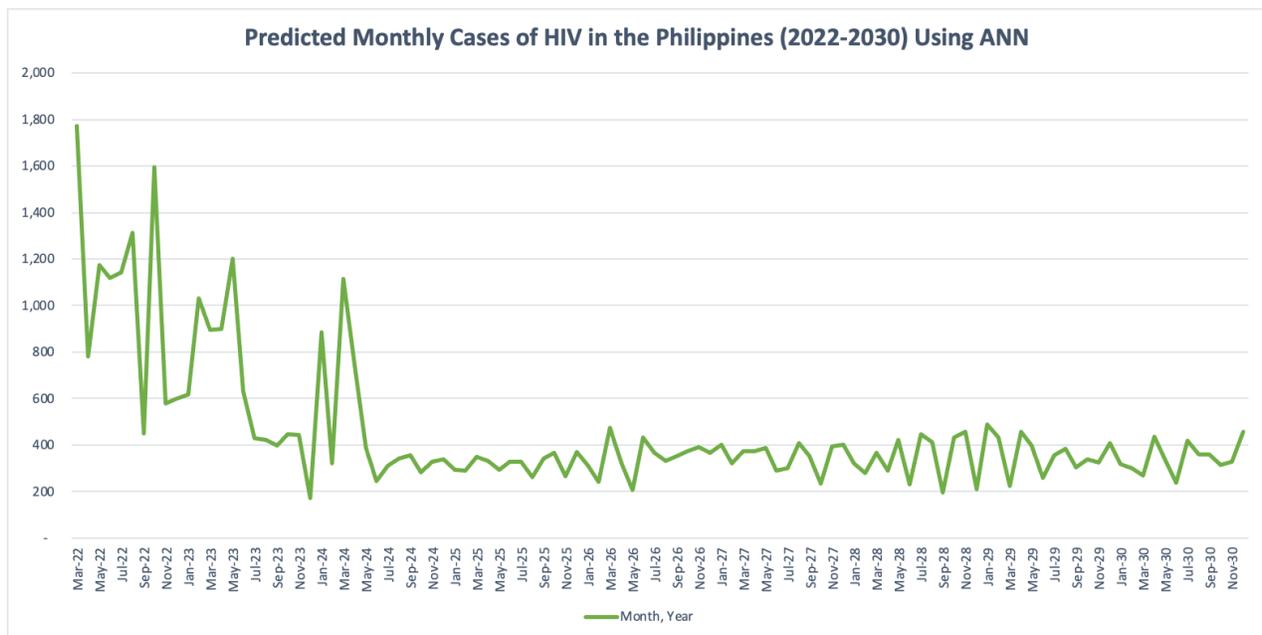

**Fig.6 Time-Series Plot of the Number of HIV Cases in the Philippines per Month**

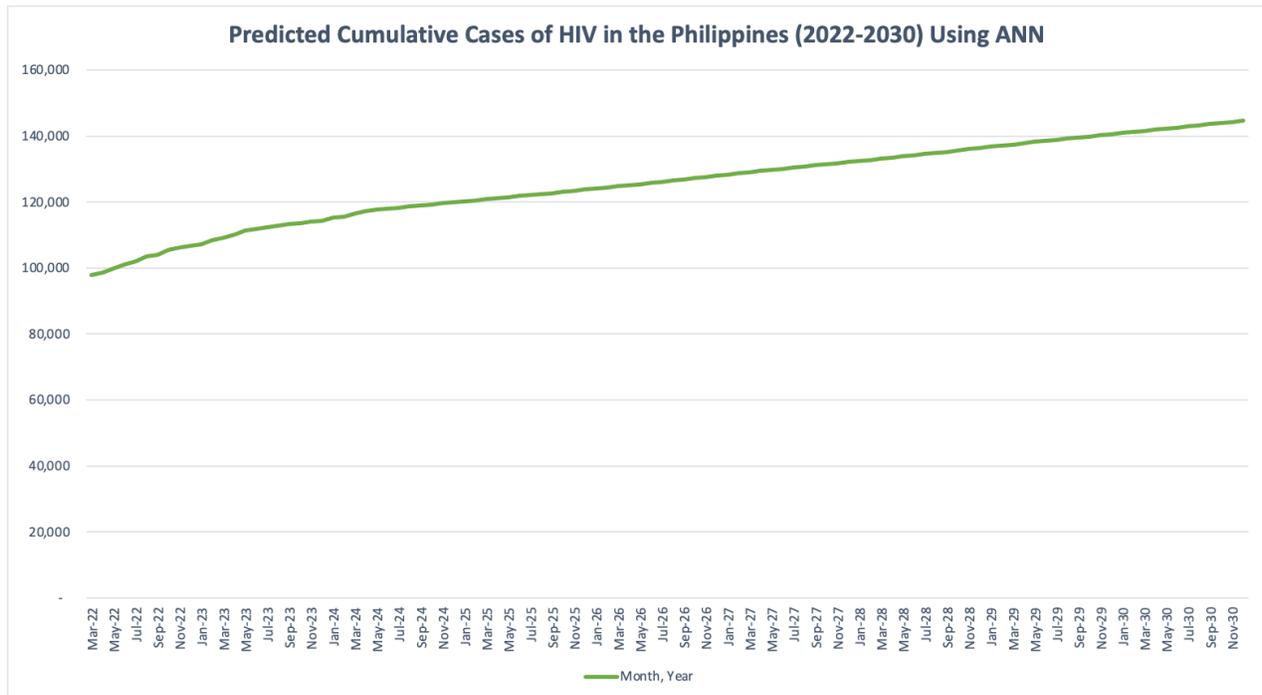

**Fig. 7 Future Estimates of HIV Aggregated Cases in the Philippines**

*C. On the Status of the Philippines' Achievements in the Implementation of SDG-3*

Without attending to the needs of those who are HIV positive, specifically on the criteria of vulnerability and health, the AIDS epidemic cannot be eliminated. The virus must be eradicated by the year 2030, according to the HIV/AIDS advocacy group Joint United Nations Programme on HIV/AIDS (UNAIDS)[50]. The General Assembly of the UN talked about this at length with a commitment to stop the HIV epidemic by 2030, otherwise known as Project 2030's SDG-3 [51]. Thus, the quantifiable evaluation of this project's achievement is a 90% decline in new HIV occurrences between 2010 and 2030[52].

The Philippines reported 174 confirmed new infections in December 2010. However, the ANN forecasting model predicts that by December 2030, there will be 457 new cases of HIV, a 162.91 percent rise. Additionally, as per the yearly case, the 2010 statistics showed a new HIV incidence of 1,591, whereas the 2030 annual estimate forecasts a case increase to 4,144, a 160.49 percent growth.

Therefore, despite the strict HIV testing and treatment cascade, the Philippines is still far from putting an end to the HIV epidemic.

*D. On Assessing the Difference between the Actual and Forecasted HIV Epidemic Values*

The comparison between observed and predicted HIV/AIDS per month during the neural network training, validation, and testing is shown in Figures 8 and 9. The actual HIV cases and the predictions generated by MATLAB software were plotted using Microsoft Excel to show comparison through chart generation. The figure depicts that although there have been some minor fluctuations in some situations, the projected and actual HIV cases follow a similar pattern. The outcome also suggests that most of the anticipated and actual HIV cases are comparable and identical in all circumstances. The overall result is therefore primarily considered to be highly significant.

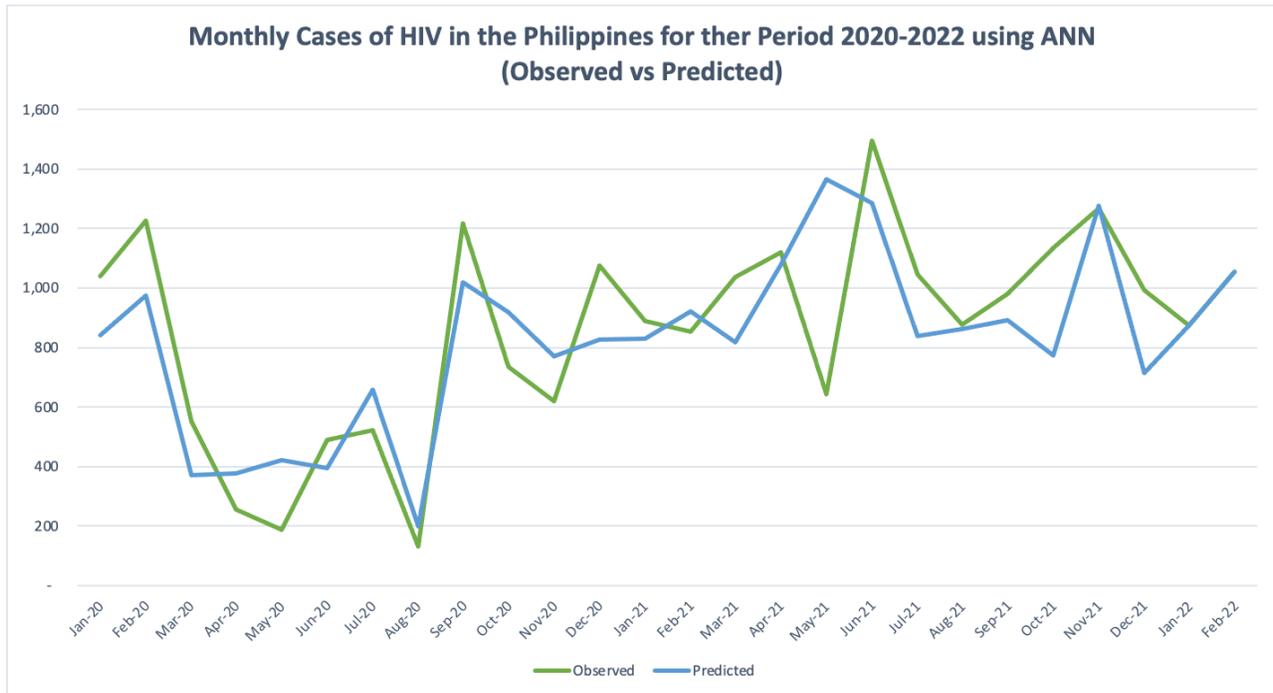

**Fig. 8 Monthly Comparison of HIV Cases in the Philippines (Actual versus Forecast)**

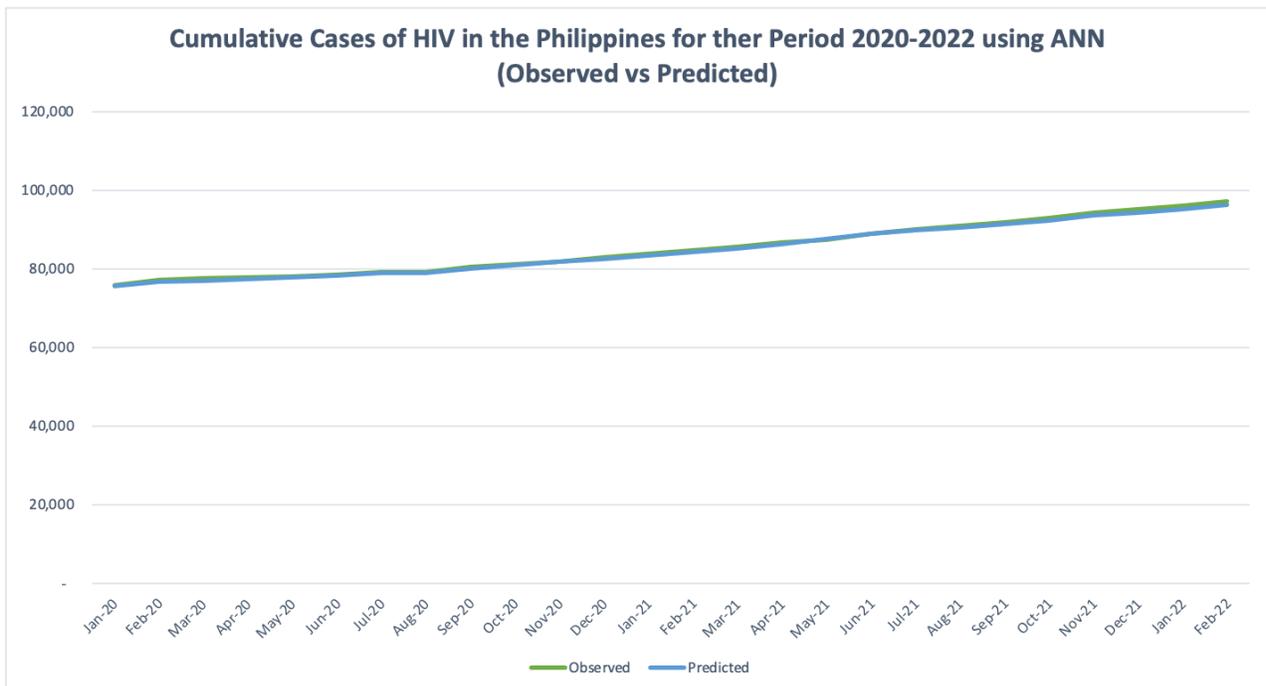

**Fig. 9 Comparison of the Aggregated HIV Cases in the Philippines (Actual versus Forecast)**

*E. On the ANN Forecasting Model's Evaluation Metrics and Performance Results*

The error and accuracy rate of the model for the trained and tested data are displayed in Table 5. The procedure was successful, as shown by the meager numbers that the RMSE, MAPE, and MAE output produced. Furthermore, the COVID-19 sets of data used to generate a Pearson's $R^2$ equivalent to 0.5872 suggest a nearly perfect fit and, therefore, a highly dependable model for this specific forecast based on Model Goodness-of-Fit.

TABLE 5
PERFORMANCE RESULTS OF THE TRAINED DATA

| Evaluation Metrics | Results |
|---|---|
| RMSE | 36.9200 |
| MAE | 180.7600 |
| MAPE | 28.5400 |
| $R^2$ | 0.5872 |

### IV. CONCLUSIONS AND RECOMMENDATIONS

From the findings reported, the following conclusions can be drawn:

1. It has been determined that the COVID-19 data sets can be used to predict future HIV cases in addition to being utilized for training, validating, and testing the model.
2. In December 2030, there will be 457 monthly infections and 145, 273 aggregated cases in the nation, indicating an upward trend line with a cycle and periodicity. Additionally, the monthly trends exhibit irregular cycles that slide downward in the later years.
3. Despite strict HIV diagnostic and therapeutic programs, the Philippines seems far from eliminating the HIV pandemic following SDG-3 mandates due to a rise in HIV infection rates, notably both monthly (162.91%) and annually (160.49%).
4. Since the actual and predicted HIV case values are similar and comparable, the overall result is highly regarded as accurate and valid.
5. The suggested model produces credible and correct outcomes based on evaluation metrics and performance results.

The following suggestions could be made based on the study's findings and conclusions:

1. The DOH and World Health Organization (WHO) may use these findings to improve their HIV prevention, testing, treatment, and care program because COVID-19 worsens access to HIV diagnosis and viral load suppression while decreasing access to HIV prevention services and testing, which could deteriorate HIV epidemic control.
2. The researcher recommends using optimizers such as Particle Swarm Optimization (PSO) to improve the performance of the existing model.